\documentclass[sigconf]{acmart}
\AtBeginDocument{%
  }

\acmConference[]{Recsys '25: CONSEQUENCES Workshop}{September, 2025}{Prague}

\usepackage{amsmath}
\usepackage{listings}
\usepackage{graphicx} 
\DeclareMathOperator*{\argmax}{arg\,max}
\usepackage{makecell}

\newcommand{\E}{\mathbb{E}}
\renewcommand{\P}{\mathbb{P}}

\begin{document}

\title{Offline Contextual Bandit with Counterfactual Sample Identification}

\author{Alexandre Gilotte}
\email{a.gilotte@criteo.com}
\orcid{0009-0003-7144-934X}
\affiliation{%
  \institution{Criteo AI Lab }
  \city{Paris}
  \country{France}
}
\author{Otmane Sakhi}
\email{o.sakhi@criteo.com}
\affiliation{%
  \institution{Criteo AI Lab }
  \city{Paris}
  \country{France}
}
\author{Imad Aouali}
\email{i.aouali@criteo.com}
\affiliation{%
  \institution{Criteo AI Lab and CREST-ENSAE}
  \city{Paris}
  \country{France}
}

\author{Benjamin Heymann}
\email{b.heymann@criteo.com}
\orcid{0000-0002-0318-5333}
\affiliation{%
  \institution{Criteo AI Lab, Fairplay joint team}
  \city{Paris}
  \country{France}
}


\renewcommand{\shortauthors}{Gilotte et al.}

\begin{abstract}

 In production systems, contextual bandit approaches often rely on direct reward models that take both action and context as input. However, these models can suffer from confounding, making it difficult to isolate the effect of the action from that of the context. We present \emph{Counterfactual Sample Identification}, a new approach that re-frames the problem: rather than predicting reward, it learns to recognize which action led to a successful (binary) outcome by comparing it to a counterfactual action sampled from the logging policy under the same context. The method is theoretically grounded and consistently outperforms direct models in both synthetic experiments and real-world deployments.
\end{abstract}







\maketitle

\section{Introduction}
\textbf{Contextual Bandits}
\cite{lattimore19bandit} serve as an important intermediate framework between multi-armed bandits and full reinforcement learning (RL) \cite{sutton98reinforcement}. Like RL, they enable decision-making based on rich, high-dimensional state (or context) information. However, they simplify the problem by assuming that contexts are independent and identically distributed (\emph{i.i.d.}).
Because many real life systems, such as some recommender systems \cite{recsys_mips}, are well approximated by contextual bandits, there is value in designing practical algorithms for the Contextual Bandit model. 
A commonly used framework is the \textbf{offline Contextual Bandit}~\cite{swaminathan2015batch}, in which a dataset has been collected by an initial policy, and the goal is to learn from this dataset a policy with the best expected reward.

\paragraph{IPS based methods}
The best performing methods for offline contextual bandits are usually building an estimator of the expected reward of a policy by \textit{Inverse Propensity Scoring }(IPS)~\cite{bottou2013counterfactual,horvitz1952generalization, dudik2011doubly}, and searching for a parametrized policy directly maximizing this criteria. While simple, IPS implementations may suffer from high variance, many improvements have been proposed to make it more stable~\cite{swaminathan2015batch,london2019bayesian,zhu2022contextual,jeunen2021pessimistic,sakhi2023pac,sakhi2024logarithmicsmoothingpessimisticoffpolicy,aouali23a,aouali2024unified,su2019cab,su2020doubly,metelli2021subgaussian,kuzborskij2021confident}.

\paragraph{Direct Method}
\label{sec:direct_metod}
In practice,
a simple, commonly used algorithm for contextual bandit consists in fitting a model of the expected reward, as a function of the context $x$ and the chosen action $a$.
This is done by  applying a supervised learning algorithm  on the available dataset \citep{sakhi2020blob,aouali2024bayesian}. 
Then, from this model, a greedy policy is built by returning the action which maximize the model-estimated expected value.
This method is sometimes referred as \textit{Direct Method} (DM), or \textit{Q-learning} in the RL settings \cite{sutton98reinforcement}.
While it seems that a well-tuned modern IPS algorithm would usually get better results, the Direct Method seems to be still widely used in practical settings.
There are several reasons for this popularity: 
(a) It is relatively simpler, only requiring a "classical" supervised learning algorithm.
(b) It might be more stable than IPS, or at least some IPS variants, because of the potentially high variance of the IPS estimator.
(c) It may be useful to tune the level of exploration, required for learning the next iterations of the policy, independently from the learning of the policy. While $\epsilon$-greedy exploration is one approach, practitioners often favor increased exploration of actions that the direct model estimates as near-optimal. However, since IPS methods directly output policies, it makes it  less straightforward to tune this level of exploration independently.
(d) Contextual bandits are only approximations of real systems, and residual sequence effects (e.g. the action of recommending an item to a user might slightly change the state and reward later in the sequence) make the \textit{i.i.d.} assumption not strictly true. Even in the case when these effects are small, they break the unbiasedness assumption which was a big selling point from IPS. It is unclear how IPS - or other methods - perform in such cases.\footnote{we believe that further research would be useful here. The same question applies to the method we propose in this paper.
}

\paragraph{When the Direct Method may fail to produce a good policy}
One typical case when DM may dramatically fail is when a variable which explained the action chosen in the past data is missing from the direct reward model; possibly leading to an instance of Simpson's paradox~\cite{peters2017elements}.
While in practice the past policy is usually another instance of the same model, there are still frequent cases when this may happen, e.g. if an A/B test  adds or removes some features from the model.
Another issue is that even with all relevant features available to the model, it may be non trivial to entangle the causal effect of an action from the effect of a context. This is especially the case because actions tend to be strongly correlated to contexts: both because some actions might be available only in some contexts, and because the past policy correlates actions and contexts.
Because of these correlations, regularization methods typically used on supervised learning may introduce some kind of confounding in the final model \cite{hahn2016regularizationconfoundinglinearregression}.
These issues can be made more acute on systems in which the context is much more predictive of the reward than the action. This is  typical in online advertising recommender systems, where the probability of a click on an ad may vary by 2 orders of magnitude with the inventory and user state, while between a "good" and "average" recommendation the same probability of click would typically change by a factor less than 2.

\paragraph{Ranking models} In recommendation settings, where several products are displayed together in the same context, another popular variant is the use of a \textit{ ranking loss} explicitly comparing the reward of products from the same context. This avoids modeling the effect of the context, and  usually outperform \textit{pointwise} DM.

\paragraph{Contribution} In this paper we propose a new method to learn a policy from offline data \textbf{which is halfway between IPS and DM}, Counterfactual sample identification (CSI). Like IPS, it requires the knowledge of the logging policy, and uses it to lessen the effect of confounding from context features. But like DM, it  only fits a supervised model with regular supervised learning methods. It can be interpreted as an adaptation of ranking models to the pure contextual bandit setting, while also adapting the idea of Retrospective Estimation \cite{Goldenberg_2020} to a non-binary action space.

\section{Counterfactual sample identification (CSI)}

\subsection {Notations and context}
The system receives a \textit{context} $X$, sampled from an unknown distribution. Each context comes with a finite set $\mathcal{A}(x)$ of available \textit{actions}, and the system must select one action $a$ in this set. A reward $Y$, that depends on  $x$ and $a$, is then received. We  restrict  to binary rewards, that is,  $Y$ is a Bernoulli variable whose parameter depends on $x$ and $a$.

A policy $\pi(\cdot \mid x)$ is a conditional probability distribution over actions. It maps each context $x$ to a distribution on $\mathcal{A}(x)$. The offline contextual bandit model assumes that we have a dataset made of i.i.d. samples $\mathcal{D}=(x_i,a_i,y_i)_{i\in[n]}$, where actions were sampled from a known policy $\pi_0$, and the
 goal is to choose a policy $\pi_\theta$ in the parametrized family $\{\pi_{\theta},\theta\in\Theta\}$ that maximizes the expected reward $ \E_{x} \E_{a \sim \pi_\theta(\cdot \mid x)} [Y] $.

\subsection {Method presentation}
\label{section_counterfactual_sample}

We present here 'Counterfactual Sample Identification' (\textsc{CSI}), a method to address the confounding effect from context variables. We first process the dataset as follows: (a) we keep only the positive data point from $\mathcal{D}$; (b) for each positive example $(x,a,y=1)$, we sample a "counterfactual" action $a'$ from the logging policy $a' \sim \pi_0(\cdot \mid x)$; and produce two examples $(x,a,z=1)$ and $(x,a,z=0)$  (c) we combine this two examples $(x_i,a', z=0)$ and $(x_i,a ,z=1)$ into a new log. 
Here $Z$ is an auxiliary variable indicating wether the sample contains the true action $a$ or the resample $a'$.
Said otherwise, we \emph{build}:
\begin{align*}
    \hat{\mathcal{D}}=\{(x,a',z=0):(x,a,1)\in\mathcal{D}\}\cup \{(x,a,z=1):(x,a,1)\in\mathcal{D}\},
\end{align*}
where $a'$ is sampled from $\pi_0(\cdot \mid x)$.
\noindent We then \emph{train} a supervised learning algorithm on $\hat{\mathcal{D}}$ using $z$ as target and $(x,a)$ as features. 
Let $f(x,a)$ be the resulting model.
The greedy CSI policy consists in playing $ a^*:= \argmax_{a \in \mathcal{A}(x)} f(x,a) $.
\footnote{Just as in the classical reward model case, we need to adapt it to include some exploration - We do not delve into this topic here as it is not different.}

\noindent Intuitively, the CSI model learns to recognize if the action is the "true" action which led to a reward $y_i=1$ , or just a resample from $\pi_0$ independent from $Y$. This idea is formalized in Lemma~\ref{lemma_1}, which relates the learned probability $f(x,a)$ and the expected reward $\P( Y=1 | ,A=a, X=x)$ of action $a$ in context $x$.  
\begin{lemma}
\label{lemma_1}
Let $\sigma: x \rightarrow 1/{1+\exp(-x)}$ be the sigmoid function. Under a uniform coverage policy $\pi_0$, we have the following identity:
   \begin{align*}
        \P(Z=1|A=a,X=x )=  \sigma \Bigg( \log\left( \frac{ \P( Y=1 | A=a, X=x)}{\P( Y=1 | X=x)} \right)\Bigg).
   \end{align*}
\end{lemma}

The proof can be found in Appendix~\ref{app:proof}. Note that $ \P( Y=1 | X=x)$ depends on the policy $\pi_0$, but not on $a$, this implies that $ \P(Z=1|A=a,X=a )$ orders the actions by their expected reward in context x. The fraction $ \frac {\P( Y=1 | ,A=a, X=x)} {\P( Y=1 | X=x)} $ which appears in the log in the equation above can be thought as the \emph{multiplicative advantage} of playing action $a$ in context $x$ - compared to the typical outcome when following $\pi_0$.
We therefore have
\begin{align*}
    \argmax_{a\in\mathcal{A}(x)}\underbrace{\P( Z=1 | X=x,A=a, Y=1 )}_{\approx f(x,a)} = 
        \argmax_{a\in\mathcal{A}(x)}\P( Y=1 | X=x,A=a).
\end{align*}




In a nutshell, instead of modeling the  expected reward $\P(Y|x,a)$, our method  directly models the multiplicative advantage. 
It thus avoids the necessity to learn the direct impact of the context on the reward: $P(Z = 1 | Y = 1  ,X =x ) = 0.5$  for all $x$, and thus a feature of $x$ matters in this model only if the relative performances of the actions change with this feature. It is therefore reasonable to expect that \textsc{CSI}  performs better on instances where modeling the multiplicative advantage is easier than modeling the reward.


\paragraph{Dependency on $\pi_0$} Like the importance-weight based methods, the \textsc{CSI} relies on having a dataset collected from a stochastic policy $\pi_0$ that explores well the action space, and it degenerates when this assumption does not hold.
To see why, let us assume that for a given context $x_i$ the policy is deterministic.
Then with probability $1$ both the true action $a_i$ and the resampled action $b_i$ will be identical. There is no point in trying to learn to distinguish them, the model can only learn to output "0.5" for these samples.
\paragraph{Replacing the sampling of A' by an expectation }
The sampling of $A'$ adds a source of noise in the training, which can be avoided: instead of producing one single counterfactual sample $x,a',z=0$ with a sampled $a'$, we can return one for each possible action $a$, and weight these samples in the learning by their sampling probability $\pi_0(a' \mid x)$. This does not change the expectation of the loss (we replaced the loss on samples of $a'$ by the expected loss), but reduces the variance; at the cost of a larger training set size. Empirical experiments in Section \ref{sec:xps_synthetic_data} confirm that this improves the quality of the learned policy.

\section{Empirical results}

\subsection{On synthetic data}
\label{sec:xps_synthetic_data}
We now compare, in a synthetic environment,  several offline learning methods for contextual bandits. 
Our experiments can be reproduced with a notebook shared in the supplementary material, and the description of the datasets is available in the appendix.

On these synthetic datasets we trained: A direct model of the reward (DM), a CSI model as described in  section~\ref{section_counterfactual_sample} --- either sampling the counterfactual action (CSI-sampling) or taking the expectation (CSI-expect) ) ---  and Logarithmic Smoothing, a recent differentiable IPS based method ~\cite{sakhi2024logarithmicsmoothingpessimisticoffpolicy}.

Table~\ref{tab:table_results_python} reports the mean normalized reward on 100 environments with different sample size of the collected datasets. 
With lowest sample counts, DM method performs well. However, as sample counts increase, this method seems to plateau, while the performances of CSI and IPS increase and  outperform DM. IPS gives overall the best results, as expected, but with 1M samples CSI-expect is reasonably close. 
We also note that CSI-expect consistently improves on CSI-sampling, as predicted.
Our proposed method is robust and performs well in environments where context effects are important.

\begin{table}[h!]
  \begin{center}
    \caption{Click-through rate on synthetic data }
    \label{tab:table_results_python}
    \begin{tabular}{l|c|c|c} 
      \textbf{Nb Samples} & 10K  & 100K & 500K  \\
      \hline
      DM & 0.76 & 0.83  & 0.84 \\
      CSI-sampling & 0.62  & 0.82 & 0.91 \\
      CSI-expect & 0.71 & 0.87 & 0.92 \\
      LS-IPS  & 0.82 & 0.93 & 0.96  \\
    \end{tabular}

  \end{center}
\end{table}

\subsection{On a live production system}

\paragraph{Banner design optimization}
We tried our method on our production system, selecting the \textit{design} (the size of a grid of products) of ad banners displayed on the web, on large-scale experiments, with several millions positive --- \textit{i.e.} clicked --- banners per day.

\paragraph{Production baseline} The baseline was a DM (Section \ref{sec:direct_metod}) using a logistic regression with quadratic kernel.

\paragraph{Exploration}  Online, the policy is a mixture of $5\%$ uniform exploration, with a multinomial assigning to each action a probability $\propto \hat{p}(y|x,a)^\alpha$ where $\alpha$ is a temperature parameter.


\paragraph{Learning from counterfactual samples}
We applied the (CSI), learning a model $\hat{p}(Z=1|x,a,y=1)$ with \emph{the same logistic regression}, features and second order interactions as the previous production model.
We only re-tuned the L2-regularization of this logistic.


\paragraph{Experiments with a reduced set of features}

For privacy reasons, we wanted to be able to learn a policy with only a subset of the context features which were used in the production model.
Directly fitting a reward model with only this subset of features was performing poorly. Using the method proposed here allowed to find a policy with the same subset of features whose performances were only slightly worse than the baseline with all features; \textbf{thus proving that the down-lift observed when fitting the reward model with less features was due to confounding effects}. 


\paragraph{Experiments with the full set of features}
We also noted that the proposed method improved on our production system \textbf{by more than $1\%$ according to an IPS estimate}. 
We thus tested it online, keeping the same exploration scheme as the baseline's. To ensure that the new model was able to train efficiently when gathering its own data, we split the users in 4 populations ABCD: A used the reference, trained on all available data, B used a CSI, also trained on all data, C and D used models similar to A and B; with training sets restricted to user data from their own population (so 25\% of the data).
Table \ref{tab:table_results_criteo} shows the online and offline results. \textbf{Online, policies learned from CSI  consistently over-performed the baseline  by 0.5\% to 1\%.}

\begin{table}[h!]
  \begin{center}
  \caption{Experiments on our large-scale ``banner design'' dataset. 
  Values indicate the percentage of clicks relative to the reference DM model with all features (100\%), i.e., a percentage increase or decrease compared to the reference.}    \label{tab:table_results_criteo}
    \begin{tabular}{l|c|r} 
      \textbf{Model} & \makecell{ \textbf{Clicks} \\IPS estimate } & \makecell{ \textbf{Clicks} \\Online A/B test }\\
      \hline
      DM, all features (Reference) & 100\% &   100\%\\
      DM, features subset & 94\%  & [ 92\% - 95\% ]  \\
      CSI, features subset & 99\% & [ 98.0\% - 99.0\% ] \\
      CSI, all features & 101.5\% & [ 100.4\%,100.5\% ] \\
      DM, all features, 25\% users  & 99.5\% & [99.5\%;99.7\%]  \\
      CSI, all features, 25\% users   & 100.4\% & [100.0\%;100.2\%] \\
    \end{tabular}
  \end{center}
  \vspace{-0mm}
\end{table}

\paragraph{Towards IPS learning on this system}
Since we used IPS for evaluation, and have low enough variance to use it as an estimator, why don't we directly optimize it?
We were able to learn an IPS model which confidently increased the production DM baseline; according to IPS estimator on kept out data.
However, when deployed online, the observed performances were not aligned with the expectation from IPS. We argue that it might be due to small sequences effects that were interfering with the training of this model, leading to discrepancies between offline and online performances. Also, even if this hypothesis is true, it is not completely clear why IPS was more affected by these effects than other methods. We thus believe that investigating problems that are "almost-contextual-bandit", and the robustness of different contextual bandit algorithms to such settings, is a promising future research area.

\newpage




\bibliographystyle{ACM-Reference-Format}
\bibliography{bibfile}

\appendix

\section{ Proof of the lemma \ref{lemma_1} }\label{app:proof}

We suppose that for each context $x$, $\pi_0$ covers the possible action space $\mathcal{A}(x)$, that is for each context $x$:
\begin{align*}
    \forall a \in \mathcal{A}(x),\quad \pi_0(a|x) > 0\,.
\end{align*}

Now, to clarify the definition of random variables, we note:
$A$ the action as sampled online, $A'$ the resampled action, and $B$ the action observed in the log example. That is:$$ B := \big{(} A \text{ if } Z=1 \text{ else } A'\big{)}\,.$$

For a context $x$, and any $a \in \mathcal{A}(x)$, we have:
\begin{align*}
    \P( Z{=}1 | X{=}x,B{=}a, Y{=}1 ) &=  \frac{ \P( Z{=}1 ,B{=}a, Y{=}1 | X{=}x ) }{ { \P( Z{=}0, B{=}a, Y{=}1 | x ) + \P( Z{=}1, B{=}a, Y{=}1 | X{=}x ) }} \\ 
    &= \frac{ 1}{ 1+ \frac{\P( Z=0,\, B=a,\, Y=1 \,|\, X=x )   }{ \P( Z=1, \,B=a, \,Y=1 \,|\, X=x )  } } \\
\end{align*}
By definition of $B$, when $Z=1$:
\begin{align*}
\P( Z=1, B=a, Y=1 | X=x) &=   \P( Z=1, A=a, Y=1 | X=x) \\ 
                         &= 0.5 \times \P(  Y=1 | X=x, A=a) \P(  A=a | X=x) \\
\intertext{and when Z=0, noting Y $\perp$ A':} \\
\P( Z=0, B=a, Y=1 | X=x) &=  \P( Z=0, A'=a, Y=1 | X=x)\\
                         &= 0.5 \times \P(  Y=1 | X=x) \P( A'=a | X=x)\\
                         &= 0.5 \times\P(  Y=1 | X=x) \P( A=a | X=x)\\
\end{align*}
Then:
\begin{align*}
\P( Z=1 | X=x,B=a, Y=1 )   &= \frac{ 1}{ 1+ \frac{ \P(  Y=1 | X=x) }{ \P(  Y=1 | X=x, A=a) } }\\
 &= \sigma\left( \log \left( \frac{ \P(  Y=1 | X=x, A=a) }{ \P(  Y=1 | X=x) } \right) \right) \\
\end{align*}

\section{Details on the experiments}

Experiments of Section \ref{sec:xps_synthetic_data} were run on a synthetic dataset where we have an oracle for the exact reward function and distribution of context, allowing exact evaluation of the expected reward of a policy.

\paragraph{Choice of the synthetic environment}
We wanted an environment where the effect of context is more important than the effect of the actions, to ensure that either modeling correctly or removing the effect of context is important.  Here we note that offline experiments proposed in the literature usually do not share this property: many experiments have been run on contextual bandit made from multi-label dataset, and for each context exactly one (or a few) actions lead to a reward of 1. In such settings, the best possible reward does not depend (much) on the context, we thus did not re-use these benchmarks.

\paragraph{ Description if the synthetic environment }
We run experiments on a context space $\mathcal{X} := \{0,1\}^7$ of 128 contexts described by 7 binary features; and an action space $\mathcal{A} := \{0, 1\}^5$ of 32 actions. For each run of our experiments, we started by sampling a contextual bandit environment defined as: i) a distribution on $\mathcal{X}$ and a reward model on $\mathcal{X} x\mathcal{A}$. These were defined by logistic models with random coefficients. The set of features used here to define the oracle was slightly richer than the set of features used when fitting the models or policy, to introduce some level of model miss-specification. This gives us an environment, and an oracle to estimate the expected reward of any policy.

\paragraph{ Dataset generation }
From this contextual bandit, we sampled a first dataset from an uniform policy.
We then fit a first model (either directly fitting the reward, or using the method of this paper), and collected a second dataset following the $5\%$-epsilon-greedy policy defined by this model.
We then ran different algorithm on this second dataset; and compared the results of their greedy policy.
Experiment results in the main section are average on runs where the dataset was collected following a direct model, or a CSI. In practice, we did not observe noticeable differences between the two sets of runs, but wanted to control that using an algo or another here did not change significantly the results.

\paragraph{Model details}
Both direct model and CSI in these experiments were fitted with a scikit-learn logistic regression using features from  context, action and context-action interactions: $(x, a, x \times a^T )$. The IPS model searched a policy in the same parameter space.
For each trained model, we then evaluated the greedy policy inferred from the model; normalized so that the best possible policy in this environment gets a reward of 1 and the worst possible gets 0.

\end{document}